\pdfoutput=1

\documentclass[11pt]{article}

\usepackage[final]{acl}

\usepackage{times}
\usepackage{latexsym}

\usepackage[T1]{fontenc}

\usepackage[utf8]{inputenc}

\usepackage{microtype}

\usepackage{inconsolata}

\usepackage{graphicx}

\usepackage{caption}
\usepackage{subcaption}

\title{Model Merging to Maintain Language-Only Performance in Developmentally Plausible Multimodal Models}

\author{\quad Ece Takmaz\quad Lisa Bylinina\quad Jakub Dotla\v{c}il\\
Utrecht University\\
{\tt\small  \{e.k.takmaz | e.g.bylinina | j.dotlacil\}@uu.nl}\\}

\begin{document}
\maketitle
\begin{abstract}
State-of-the-art vision-and-language models consist of many parameters and learn from enormous datasets, surpassing the amounts of linguistic data that children are exposed to as they acquire a language. This paper presents our approach to the multimodal track of the BabyLM challenge addressing this discrepancy. We develop language-only and multimodal models in low-resource settings using developmentally plausible datasets, with our multimodal models outperforming previous BabyLM baselines. One finding in the multimodal language model literature is that these models tend to underperform in \textit{language-only} tasks. Therefore, we focus on maintaining language-only abilities in multimodal models. To this end, we experiment with \textit{model merging}, where we fuse the parameters of multimodal models with those of language-only models using weighted linear interpolation. Our results corroborate the findings that multimodal models underperform in language-only benchmarks that focus on grammar, and model merging with text-only models can help alleviate this problem to some extent, while maintaining multimodal performance. 

\end{abstract}

\section{Introduction}
Current state-of-the-art multimodal language models (MLMs) are composed of many layers containing billions of parameters and they require huge amounts of data to learn how to handle and bridge visual and textual modalities. On the other hand, children acquire language with the help of much smaller sets of linguistic input. The BabyLM challenge~\cite{warstadt-etal-2023-findings} focuses on this discrepancy and encourages the implementation and training of sample-efficient, developmentally plausible models in resource-limited contexts. Although utilizing small datasets and models could prove challenging to outperform current MLMs, such setups could allow for cognitive plausibility, also making the development and use of such models more accessible and efficient.

\begin{figure}[t]
  \centering
 \includegraphics[width=1\linewidth]{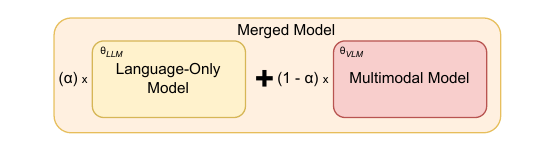}
    \caption{Weighted merging of language-only and multimodal models in the form of linear interpolation.}
    \label{fig:merged}
\end{figure}

Despite the general good performance of MLMs in multimodal tasks, previous work shows that these models tend to underperform in \textit{language-only} tasks~\cite{zhuang-etal-2024-visual}. Recent work in the multimodal BabyLM challenge also points to the same issue~\cite{amariucai-warstadt-2023-acquiring, klerings-etal-2024-developmentally}. Therefore, our aim in this paper is to first test our own models on multimodal and text-only benchmarks, and second, if we observe the same issue, to try to mitigate it. 

We develop language-only and multimodal models, the latter of which outperforms previous BabyLM baselines.\footnote{Code and models available at \url{https://github.com/ecekt/babylm_multimodal_model_merging}} However, our results indeed confirm that our developmentally plausible MLMs lack in text-only benchmarks. Hence, we explore a model augmentation technique to potentially overcome this shortcoming: \textbf{model merging}. Model merging has been utilized to prevent catastrophic forgetting and combine the capabilities of multiple models trained on different tasks, datasets, or modalities~\cite{yang2024modelmergingllmsmllms, dash2025ayavisionadvancingfrontier}.

In our approach, during inference time, we fuse the parameters of models trained on text-only and multimodal data in a straightforward, training-free way (see Figure~\ref{fig:merged}). Our results indicate that such an augmentation yields a single model maintaining accuracy and robustness in both text-only and multimodal benchmarks in the earlier and later stages of training the multimodal model.

\section{Background}
We first go into detail about the multimodal approaches to the BabyLM challenge in Section~\ref{relworkbb}, and then, provide a summary of the recent work on model merging in Section~\ref{relworkmerging}. 

\subsection{Developmentally Plausible Multimodal Language Models}
\label{relworkbb}

The BabyLM initiative encourages the development of models that can be small-scale, trained on smaller sets of data, using various techniques such as model compression, learning from interaction, knowledge distillation. Our focus is on the \textbf{multimodal track}. Multimodal models have been explored in the 2nd BabyLM challenge~\cite{choshen2024callpapers2ndbabylm} and re-introduced in the 3rd BabyLM challenge~\cite{charpentier2025babylmturns3papers} as the submitted models did not outperform the baselines released by the BabyLM organizers~\cite{hu-etal-2024-findings}. The baselines were GIT~\cite{wang2022git} and 
Flamingo~\cite{alayrac2022flamingo} models trained on the BabyLM's multimodal corpus to ground language to vision.

These models encode image inputs and generate text using text decoders conditioned on visual tokens. The methodologies from the past submissions include curriculum learning where the captions were ordered based on the number of concepts they included \cite{saha-etal-2024-exploring}, where this helped on developmentally plausible benchmarks. Pretraining on text also seems to be beneficial; \citet{saha-etal-2024-exploring} investigates first training on text and then captions along with curriculum learning using image-caption pairs. However, in general, it appears to be difficult to observe a strong pattern across model types, datasets and tasks.

Another work reports a related result where learning in phases appears to benefit multimodal BabyLM models~\cite{alkhamissi-etal-2024-dreaming}. The model first learns the language-only tasks, then grounding, followed by self-synthesized data and more advanced reasoning tasks. 

There are also contributions to the language-only track where the models were influenced or informed by multimodal input \cite{fields-etal-2023-tiny, amariucai-warstadt-2023-acquiring}.

\citet{klerings-etal-2024-developmentally} explore a weighted loss function for text-only and multimodal data during training. However, they show that vision does not significantly benefit the performance in language-only benchmarks. This is in line with prior findings showing limited or no improvements when incorporating visual data~\cite{amariucai-warstadt-2023-acquiring, zhuang-etal-2024-visual}, with the exception of low-data regimes~\citep{zhuang-etal-2024-visual}, which inspired our work.

\subsection{Model Merging}
\label{relworkmerging}
Model merging has been utilized as a technique for adaptively extending the capabilities of models or balancing performance during inference time in the tasks multiple models were trained on. See \cite{yang2024modelmergingllmsmllms} and \cite{goddard-etal-2024-arcees} for surveys of various merging techniques. 

A straightforward averaging technique called `model soups' has been found beneficial in improving accuracy and robustness. The techniques involve combining the parameters of multiple models trained on different hyperparameters, in addition to more sophisticated weighted merging methods~\cite{wortsman2022modelsoup, fisherweighted}. Similarly, \citet{aakanksha2024mix} find that merging models is better than mixing training data for facilitating safety and multilingual generalizability.

Regarding vision-and-language models, \citet{zhu2025remedy} learn modules for various multimodal tasks that are later merged; whereas ~\cite{li-etal-2025-transferring} exploit text-only reward models to transfer to vision-and-language reward models in a cross-modal model merging scheme.

Closer to our approach, AyaVision is an example of cross-modal merging to maintain text-only capabilities within multimodal models to prevent catastrophic forgetting~\cite{dash2025ayavisionadvancingfrontier}. The authors built their multimodal model on their best-performing text-only checkpoint, which makes the setup more suitable for merging. Similarly, \citet{sung-etal-2023-empirical} conduct detailed experiments on multimodal model merging, finding that simple linear interpolation is a competitive and efficient method, which we also opt for in this work to test its effectiveness in low-data and low-compute settings.

\section{Data} 
To train our models, we use the data from the multimodal BabyLM challenge, which consists of 2 parts: text-only and multimodal. 

\noindent\textbf{Text-only.} We use the 50M-word text data provided by the BabyLM challenge. This data consists of text stemming from 6 sources as explained by \citet{choshen2024callpapers2ndbabylm}, and corresponds to the first halves of the text-only subsets released for the language-only challenge of the BabyLM task. 

\noindent\textbf{Multimodal.} This part includes image-caption pairs from Localized Narratives~\cite{PontTuset_eccv2020} and Conceptual Captions 3M~\cite{sharma-etal-2018-conceptual}. We download the Localized Narratives (LN) images and captions from the dataset's website.\footnote{\url{https://google.github.io/localized-narratives/}} We use the COCO (train)~\cite{coco}
and Open Images (train and test)~\cite{openimages} subsets of LN.\footnote{Although the BabyLM OSF repository at \url{https://osf.io/ad7qg/files} provides captions and extracted image representations for this subset, we noticed a discrepancy in the number of samples compared to the original LN. It seems that the test set of the Open Images subset is also counted in the BabyLM corpus to end up with the 50M word count. Therefore, we also included that part. Localized Narrative subset IDs of Open Image downloaded from \url{https://storage.googleapis.com/openimages/web/download_v7.html}. COCO images from \url{https://cocodataset.org/}}

Additionally, we download the captions for the existing images from the Conceptual Captions (CC3M) dataset.\footnote{We download the captions from \url{https://ai.google.com/research/ConceptualCaptions/download} and the images using the script provided at \url{https://github.com/igorbrigadir/DownloadConceptualCaptions}.} Filtering out the images that do not exist anymore as well as the corrupt and duplicate image files, we end up with fewer than 3M images, which is lower than the provided captions for the previous multimodal BabyLM challenge.

\begin{table}[]
\centering \footnotesize
\begin{tabular}{lcc}
\textbf{Data}        & \textbf{Train}              & \multicolumn{1}{c}{\textbf{Val}} \\ \hline
Localized Narratives & 729349                      & 38387                             \\
CC3M                 & 2061837                     & 108518                            \\
BabyLM - text-only   & 5492930                     & 289102                            \\ \hline
Total                & \multicolumn{1}{l}{8284116} & \multicolumn{1}{l}{436007}        \\ \hline
\end{tabular}
\caption{Number of samples per dataset in the splits we created (after filtering, validating and deduplicating CC3M images and captions, and trimming it to fit 100M words in total).}
\label{stats}
\end{table}

\noindent\textbf{Final dataset.} The statistics of our final dataset are provided in Table~\ref{stats}, with our random train and validation splits where 95\% of each subset contributes to the training set and the rest goes into the validation set.

\section{Methodology}
\noindent\textbf{Models.} We modify the implementation of LLaVa~ \cite{liu2023visual, Liu_2024_CVPR} from HuggingFace\footnote{`llava-hf/llava-1.5-7b-hf'},  
inheriting the LlavaForConditionalGeneration model, and replace the visual encoder with DINOv2-large~\cite{oquab2024dinov}\footnote{`facebook/dinov2-large'}. We also make necessary changes to the image processing code and modeling code in relation to the dimensions of the image features. We randomly initialize a 6-layer version of this model as the language model, together with a mapping layer that projects the image representations to the language model's space. 

\noindent\textbf{Image representations.} Unlike the BabyLM benchmark's image representations from last year that are 768-dimensional vectors from DINOv2 ViT-Base, we use the large version of DINOv2, which processes images into 256 image tokens of 1024 dimensions. While we originally intended to feed all 256 image tokens extracted from the vision encoder, due to time and compute constraints, we modified the model to feed a single pooled image token directly. We implemented a version of the model where we pre-extract all image token representations and mean-pool them. This single summary token (1024 dim) is fed to the LM directly (bypassing the vision tower). The summary image representation goes through a multimodal projector composed of a linear layer projecting from 1024 to 768 dimensions, GeLU activation and another linear layer projecting from 768 to 768. This multimodal projector is trained along with the language model, while keeping the image representations frozen.

For text-only data, we create a black image (640 x 420) and always input the features of this placeholder image both in the text-only model and the multimodal model.

\noindent\textbf{Training the tokenizer.} Using all the text in the final dataset (token count = 100M), we train a tokenizer from scratch employing the configurations of the LLaVA tokenizer (LlamaTokenizerFast, a byte-pair encoding model based on SentencePiece), with a vocabulary size of 30000 including a special token for image representations. Using the BERT pre-tokenizer, we apply splitting on whitespace and punctuation. This preprocessing yields a 1.36 word-to-subword ratio.\footnote{Words: 99,999,990. Subwords (as tokenized by our tokenizer, skipping special tokens): 136,034,832.}

\noindent\textbf{Intermediate checkpoints.} To investigate the learning speed and model behavior dynamics, we save checkpoints (every 1M words up to 10M, every 10M up to 100M, every 100M up to 1B). We estimate the words-seen using the ratio of wordpieces to actual words in our dataset (1.36). We use this ratio to roughly determine how many `words' the models were exposed to in the training batches (excluding special tokens).

\noindent\textbf{Hyperparameters and setup.} We follow the restrictions of the BabyLM challenge, 100M words, 10 epochs, resulting in 1B tokens seen in total. 

We set the maximum length to 150 tokens. %
We truncate longer samples if they do not fit this constraint; if they are shorter, we pad them.

We train the models on 2 A10 GPUs with 24GB memory on CrossEntropy Loss using the AdamW optimizer with a learning rate of 1e-4. For the multimodal model, batches can contain text-only or multimodal data, and the loss is calculated in the same way for both modalities. We use fp16 half precision and make use of the Accelerate library for data parallelism to speed up training. We accumulate gradients for 8 steps and then apply gradient updates to optimize the model, effectively increasing our batch sizes from 64 to 512. The numbers of words seen are gathered from each GPU and only logged and checked in the main process. We opted for a smaller layer number (6) to allow for a speed-up in the training by exposing the model to larger and more batches in a shorter amount of time. It takes 6.7 days for the multimodal model (with text-only and multimodal data) to be trained, and the text-only model 4.3 days.

\section{Model Merging at Inference Time}
Since our language-only and multimodal models share the same architecture, random initialization and the text-only data, they can be combined in a straightforward way. We apply a simple weighted sum of the multimodal model's parameters and the text-only model's parameters. We experiment with merging weights $\alpha$ of 0.3, 0.5, and 0.8,\footnote{The weights were chosen to reflect equal contribution from both models (0.5) and a skewed contribution from one model (0.3--more VLM and 0.8--more LLM).} where $\theta$ indicates all the trainable parameters of a model:
\begin{equation}
    \theta_{merged} = \alpha \theta_{LLM} + (1-\alpha) \theta_{VLM}
\end{equation}

In this way, the merged model is a linear interpolation of the multimodal and text-only models.

\section{Benchmarks}
We modify the evaluation pipeline provided by the BabyLM challenge\footnote{\url{https://github.com/babylm/evaluation-pipeline-2025}} to run zero-shot evaluations across our checkpoints using the benchmarks provided. For Winoground, we write our own evaluation code.

\noindent\textbf{Language-only benchmarks.} We run the evaluation pipeline for all the tasks in BLiMP~\cite{blimp}, EWoK, entity tracking (assign the highest probability to the correct continuation)~\cite{kim-schuster-2023-entity}, Wug past tense~\cite{weissweiler-etal-2023-counting}, wug adjective nominalization~\cite{wugadj} testing morphological capabilities (correlating model probabilities to human judgments). BLiMP and BLiMP-supplement (more challenging samples) evaluate whether the models capture grammatical phenomena, where one grammatical and one ungrammatical sentence are pitted against each other, testing models' capabilities related to syntax, morphology and semantics. EWoK~\cite{ivanova2025elementsworldknowledgeewok} focuses on world knowledge and reasoning about e.g., social, physical, spatial relations. 

\begin{figure}[h]
  \centering
  \begin{subfigure}{0.43\linewidth}
    \includegraphics[width=\columnwidth]{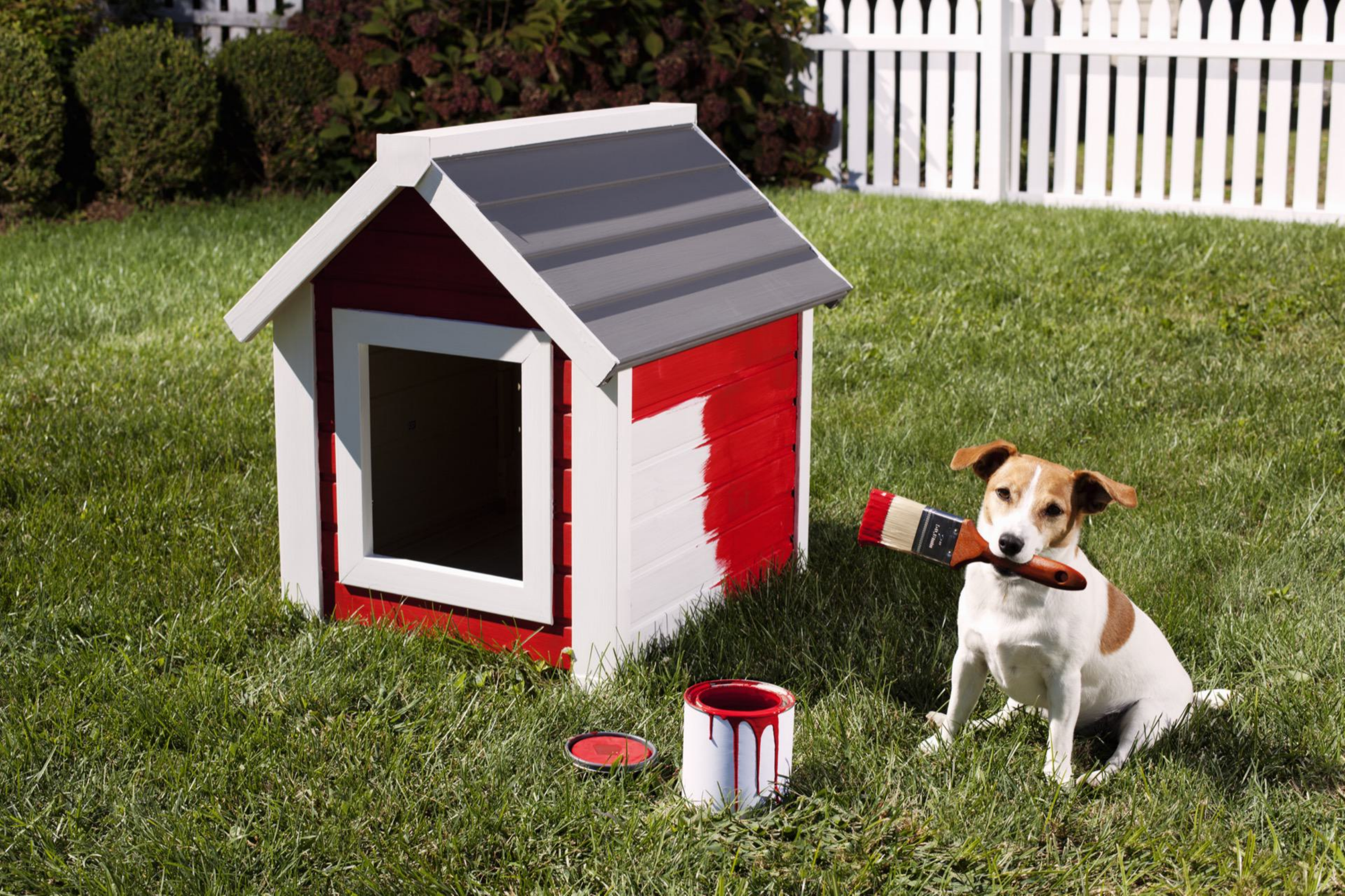}
    \label{fig:short-a}
  \end{subfigure}
  \hspace{0.2cm}
  \begin{subfigure}{0.44\linewidth}
    \includegraphics[width=\columnwidth]{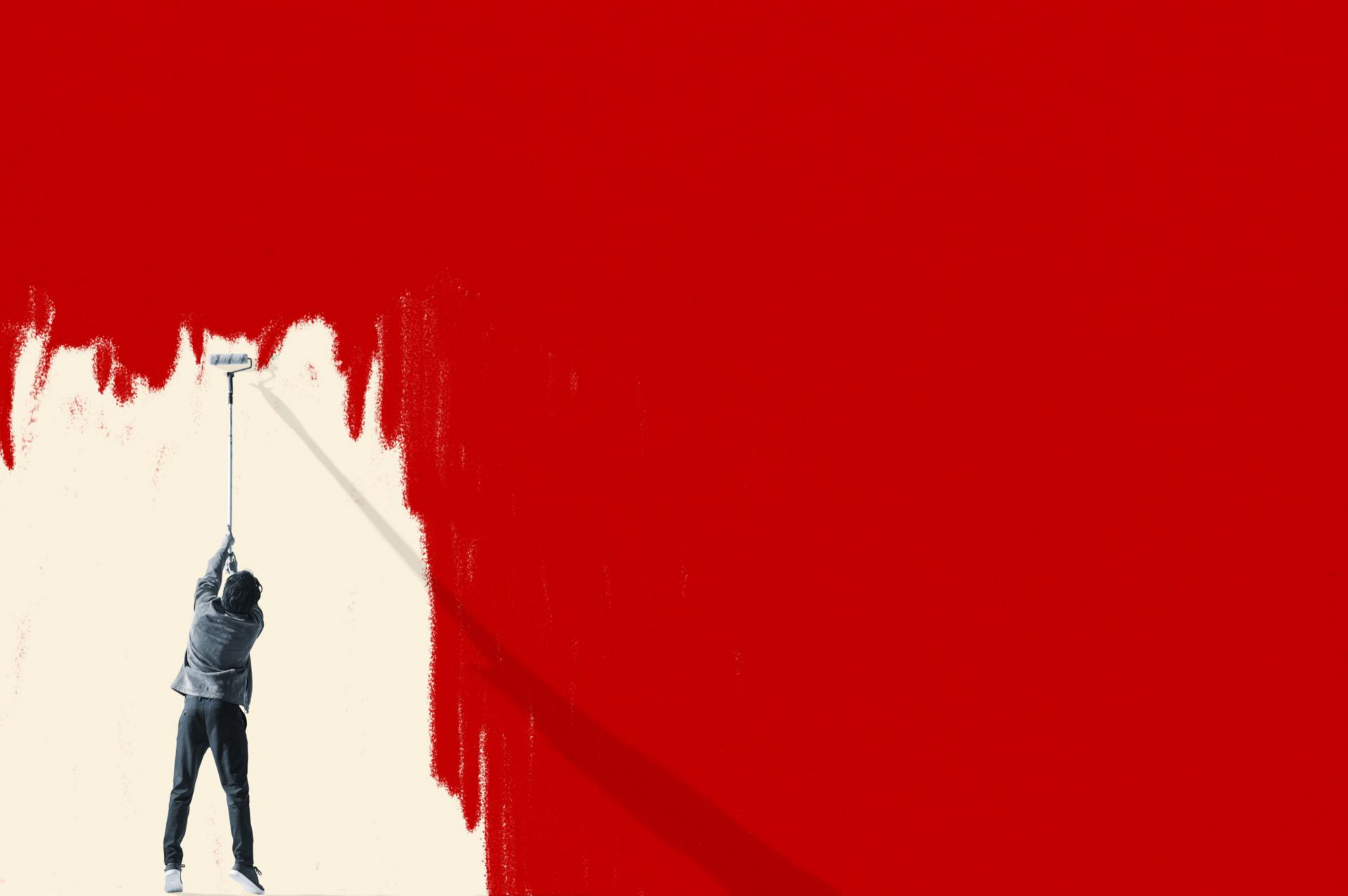}
    \label{fig:short-b}
  \end{subfigure}
  \caption{An example from Winoground. \textit{Left:} `painting the white wall red'. \textit{Right:} `painting the red wall white'. }
  \label{fig:wing}
\end{figure}

\noindent\textbf{Multimodal benchmark.} We experiment on Winoground~\cite{thrush2022winoground}, where pairs of images with very similar captions are provided. Winoground consists of 400 samples, where each sample has 2 images and 2 captions. These 2 captions have the same words, but in different orders to match the image contents (see an example in Figure~\ref{fig:wing}).

We input the image and 2 captions separately to the model to obtain predictions for Winoground. If the likelihood of the correct caption is higher, we increase the accuracy. We use the unpaired text-score as used in previous BabyLM work, where we consider each image-caption pair separately. We use the full Winoground dataset available on HuggingFace, unlike the filtered version in the BabyLM evaluation suite.

\begin{figure*}[]
  \centering
 \includegraphics[width=0.8\linewidth]{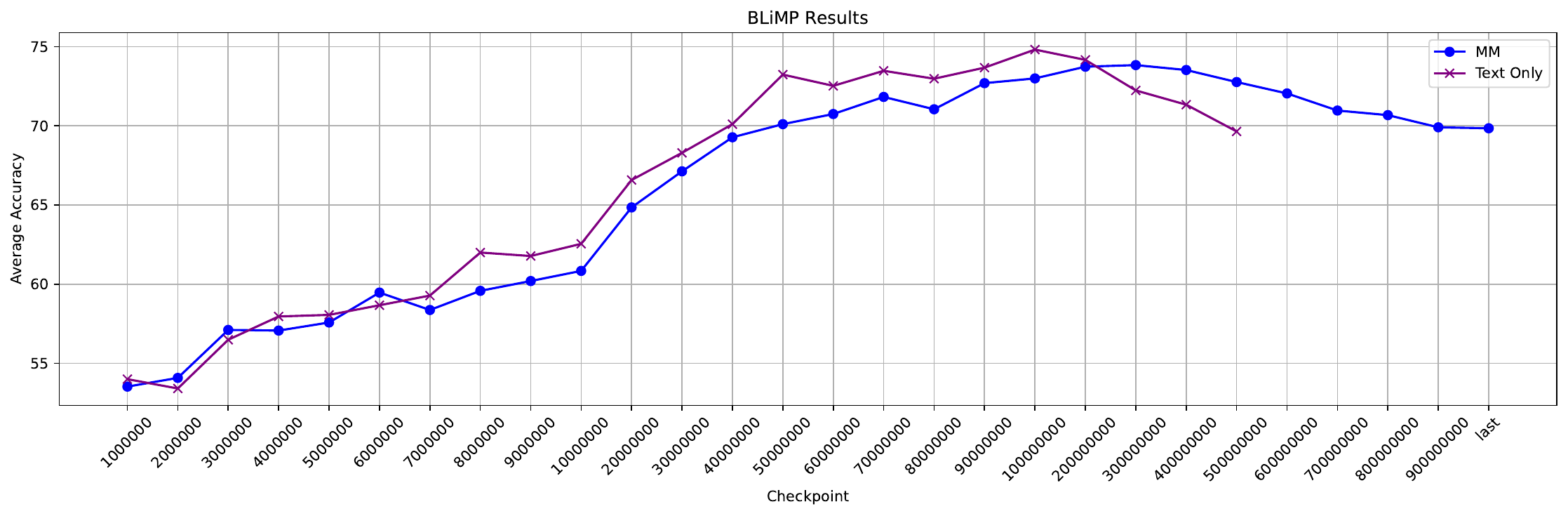}
    \caption{Average accuracies for the text-only model and the multimodal model over the training checkpoints, for the BLiMP \textbf{full} benchmark.}
    \label{fig:blimptr_full}
\end{figure*}

\begin{figure*}[]
  \centering
 \includegraphics[width=0.8\linewidth]{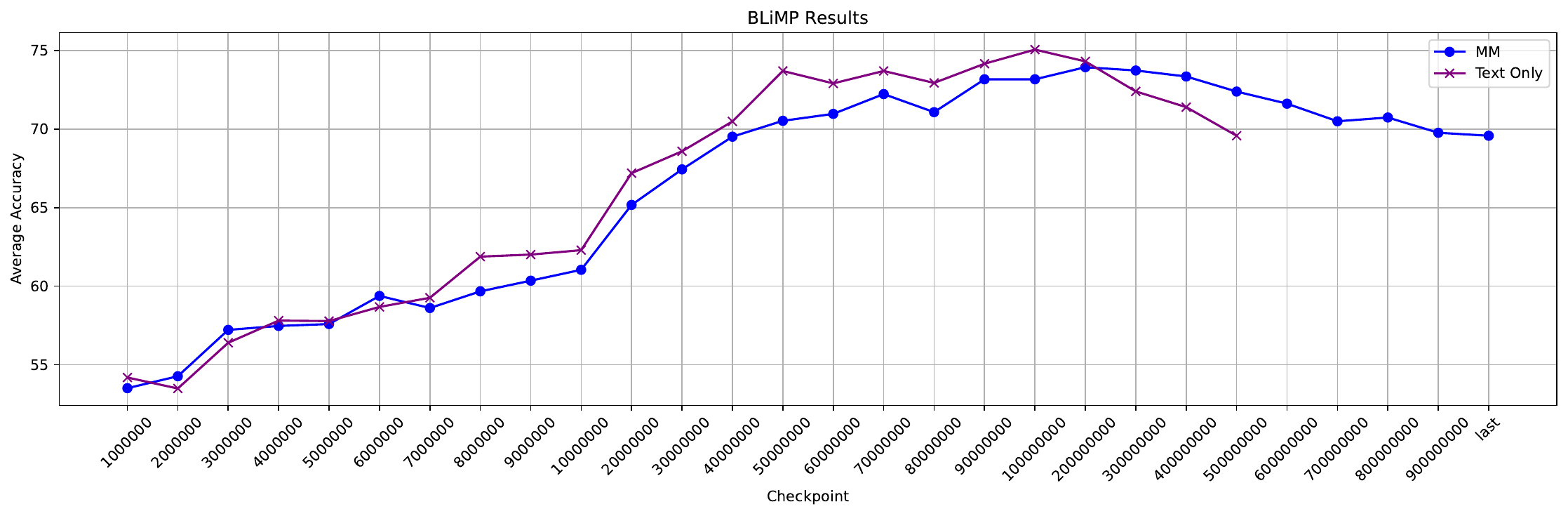}
    \caption{Average accuracies for the text-only model and the multimodal model over the training checkpoints, for the BLiMP \textbf{fast} benchmark.}
    \label{fig:blimptr_fast}
\end{figure*}

Winoground tests abilities requiring compositionality, sensitivity to word order, commonsense reasoning, pragmatics and overall more fine-grained visual and linguistic analyses involving unusual images and texts~\cite{diwan-etal-2022-winoground}. Winoground is a difficult dataset, with previous BabyLM work yielding accuracies as follows: 2024 baselines Flamingo: 51.6
GIT: 55.5; 2025 baselines Flamingo: 54.8, GIT, 56.2.\footnote{2024 baselines from: \url{https://github.com/babylm/evaluation-pipeline-2024/}, 2025 baselines from: \url{https://huggingface.co/spaces/BabyLM-community/babylm-leaderboard-2025-all-tasks}}

\begin{figure*}[]
  \centering
 \includegraphics[width=0.8\linewidth]{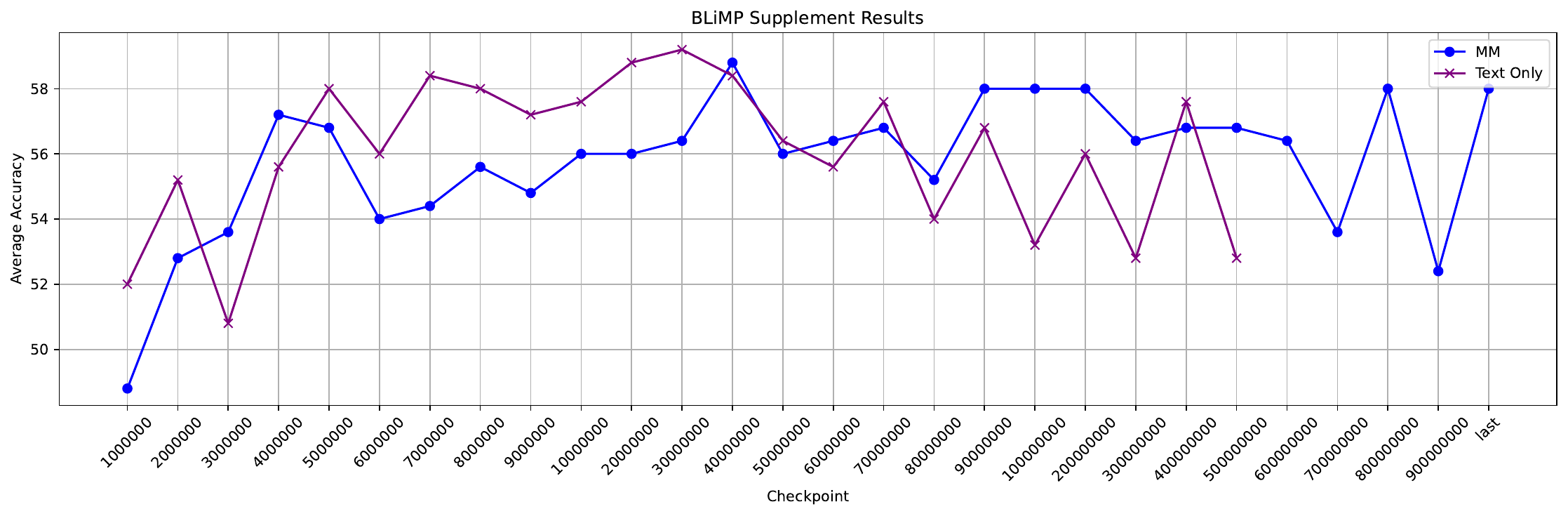}
    \caption{Average accuracies for the text-only model and the multimodal model over the training checkpoints, for the BLiMP supplement fast benchmark.}
    \label{fig:suptr}
\end{figure*}

\begin{figure*}[]
  \centering
 \includegraphics[width=0.8\linewidth]{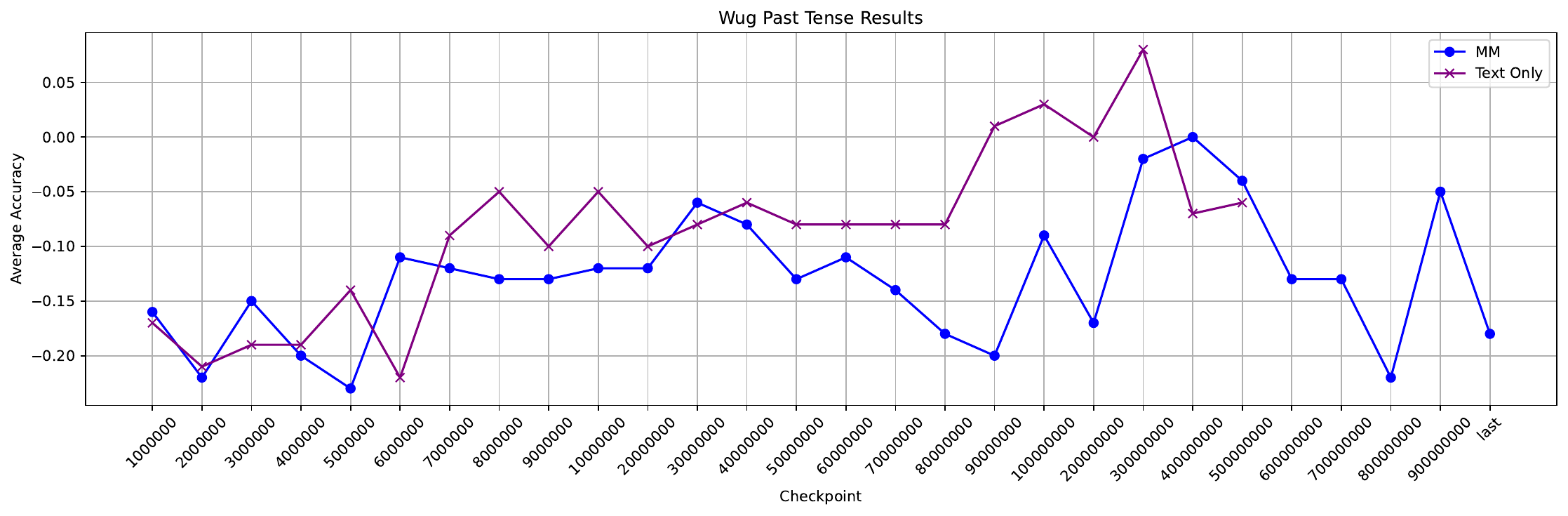}
    \caption{Correlation between model predictions and human responses from the Wug past tense benchmark, for the text-only model and the multimodal model over the training checkpoints.}
    \label{fig:wugpst}
\end{figure*}

\begin{figure*}[]
  \centering
 \includegraphics[width=0.8\linewidth]{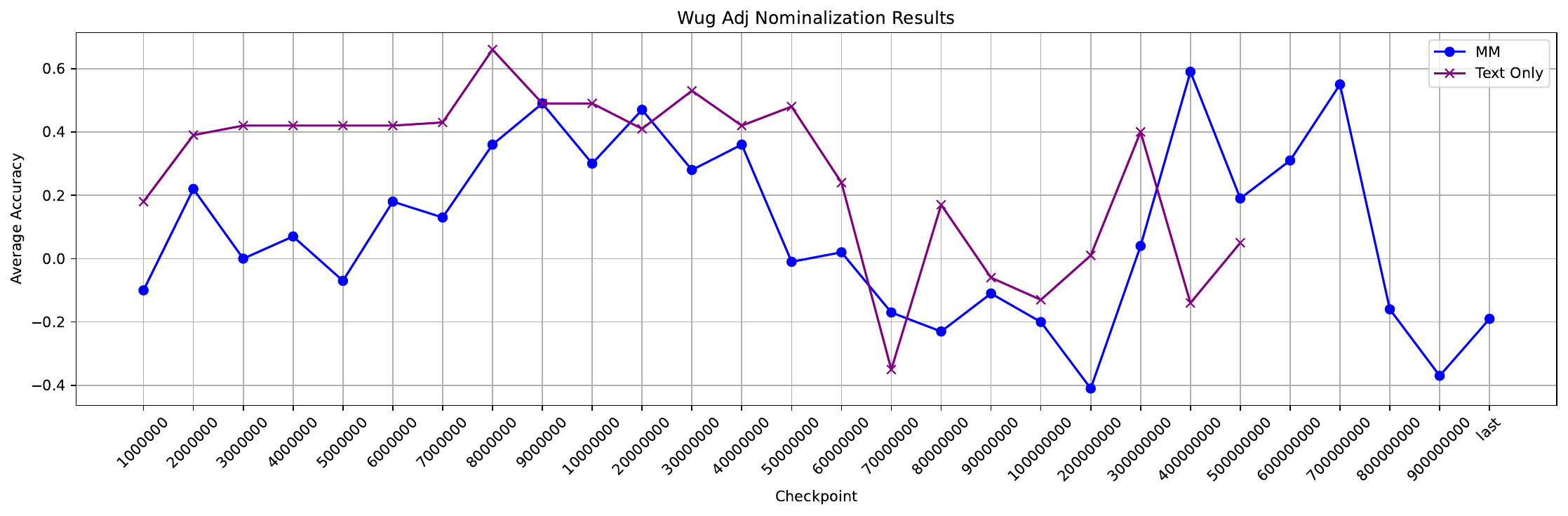}
    \caption{Correlation between model predictions and human responses from the Wug adjective nominalization benchmark, for the text-only model and the multimodal model over the training checkpoints.}
    \label{fig:wugan}
\end{figure*}

\begin{figure*}[]
  \centering
 \includegraphics[width=0.8\linewidth]{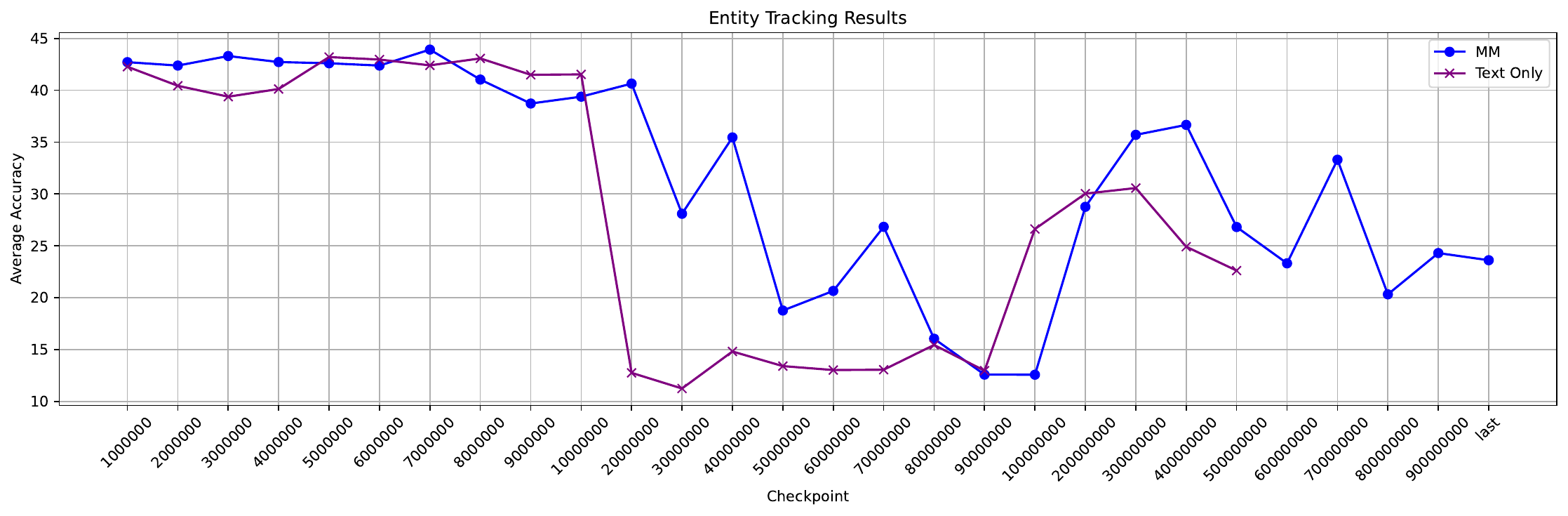}
    \caption{Average accuracies for the text-only model and the multimodal model over the training checkpoints, for the Entity Tracking fast benchmark.}
    \label{fig:enttr}
\end{figure*}

\begin{figure*}[]
  \centering
 \includegraphics[width=0.8\linewidth]{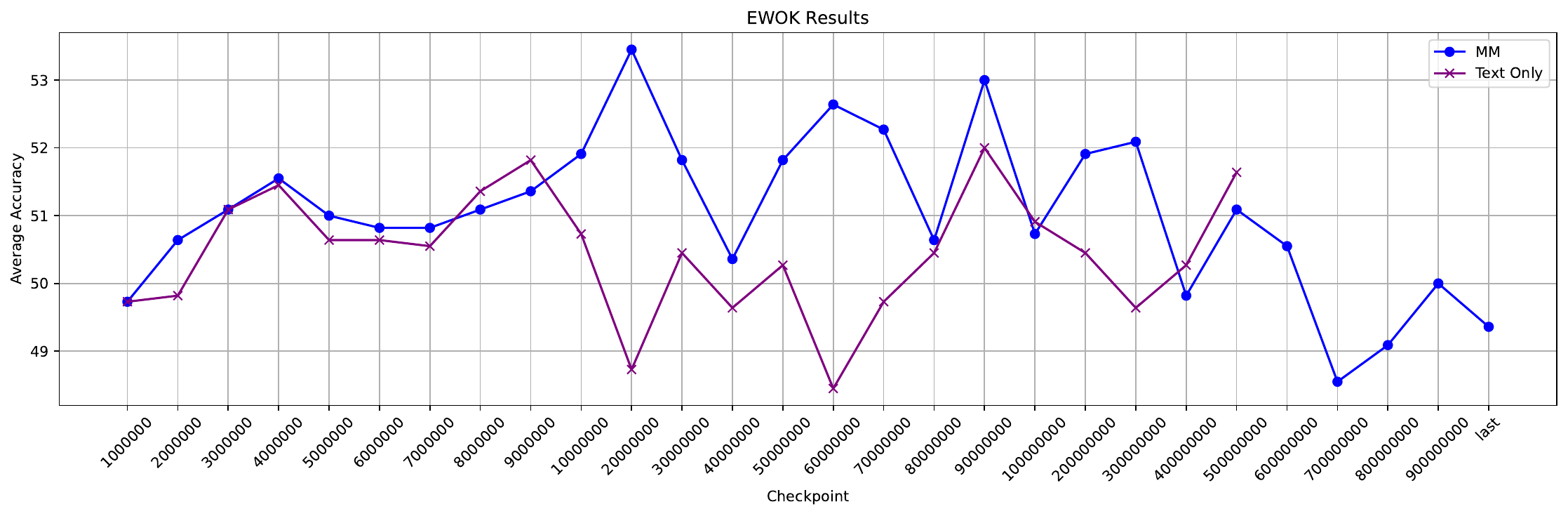}
    \caption{Average accuracies for the text-only model and the multimodal model over the training checkpoints, for the EWoK fast benchmark.}
    \label{fig:ewok}
\end{figure*}

\begin{figure*}[]
  \centering
 \includegraphics[width=0.8\linewidth]{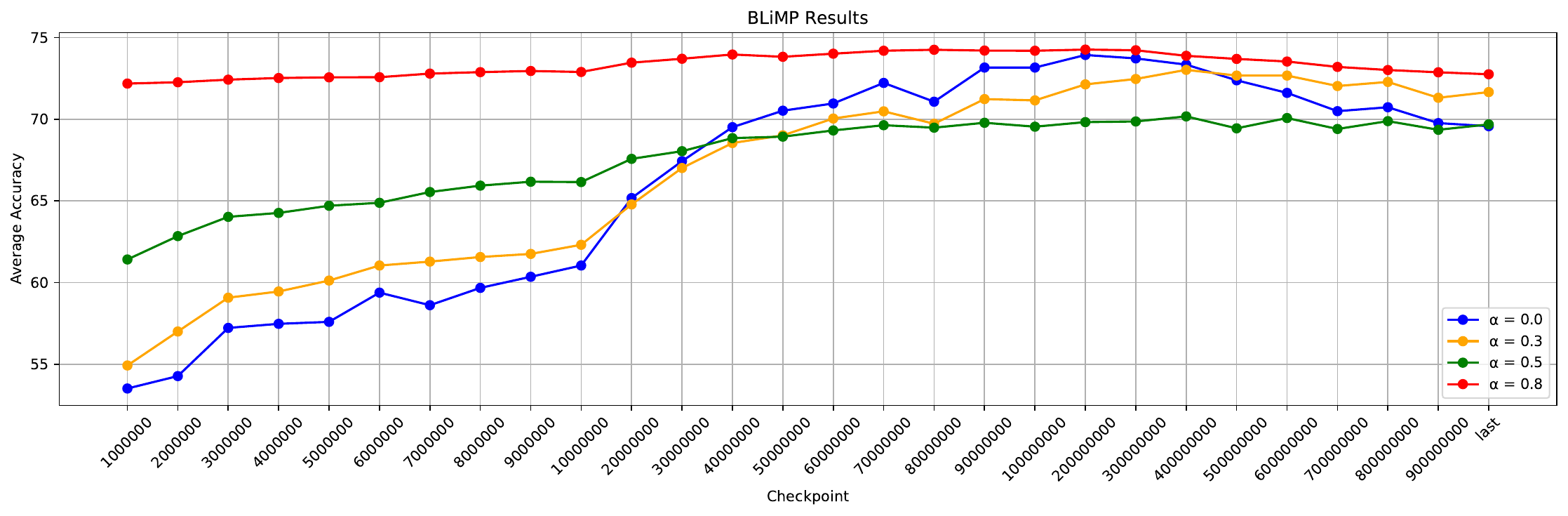}
    \caption{Average accuracies for the merged models with different weights (higher $\alpha$ indicates more contributions from the language-only model), along with the training dynamics of the multimodal model for the BLiMP fast benchmark.}
    \label{fig:mergeblimp}
\end{figure*}

\begin{figure*}[]
  \centering
 \includegraphics[width=0.8\linewidth]{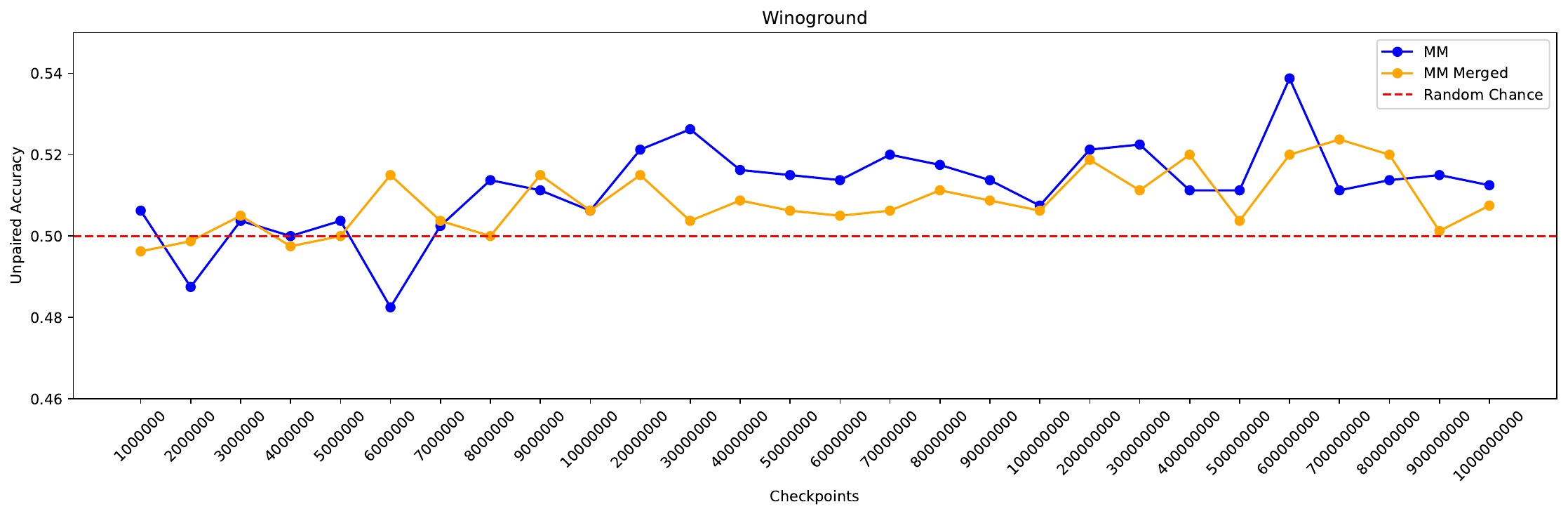}
    \caption{Average accuracies for Winoground. MM represents the multimodal model checkpoints and MM Merged indicates a merged model with $\alpha=0.3$, using the language-only checkpoint with the highest BLiMP score. The red dotted line indicates random chance accuracy. The top score in the leaderboard on the filtered version is 56.2.}
    \label{fig:wgmerged}
\end{figure*}
\section{Results}

\noindent\textbf{Results on benchmarks.}
We first obtain the results on the full BLiMP evaluation, which is reported in Figure~\ref{fig:blimptr_full}. Our best language-only model reaches 74.82 accuracy at the 100M checkpoint. \textbf{Our best multimodal model yields 73.84 accuracy, surpassing the multimodal BabyLM baselines as well as the current 2 submissions on the multimodal BabyLM leaderboard} (2024 Flamingo: 70.9, GIT: 65.2, 2025 Flamingo: 70.9, GIT: 72.2). We see that, generally, the text-only model outperforms or is on par with the multimodal model, except some later checkpoints.

We use the `fast' versions of the benchmarks that contain a smaller set of samples to obtain the following results due to time and compute constraints.\footnote{We noticed that the Wug fast and full benchmarks are in fact identical.} In Figure~\ref{fig:blimptr_fast}, we depict the performance of the text-only and multimodal model checkpoints on the BLiMP fast benchmark, which yields outcomes closely aligned with those obtained from the full benchmark.

We see similar trends for BLiMP supplement (Figure~\ref{fig:suptr}), Wug past tense (Figure~\ref{fig:wugpst}) and adjective nominalization (Figure~\ref{fig:wugan}) benchmarks. This is in line with previous work indicating that multimodal data tend not to benefit performance on language-only benchmarks. 

When we look at the results on the Entity Tracking (Figure~\ref{fig:enttr}) and EWoK benchmarks (Figure~\ref{fig:ewok}), however, we see trends where multimodal checkpoints clearly outperform the text-only checkpoints. This could be due to the focus of these datasets, which is more knowledge- and semantics-oriented rather than grammatical, therefore, the multimodal data such as the image descriptions in narrative form from the Localized Narratives dataset could have helped. 

Although our models do not perform well in the BLiMP supplement and Wug past tense benchmarks, they show competitive performance in the remaining tasks.

\noindent\textbf{Results on model merging.} We use BLiMP as a use case for our model merging experiments. We merge each multimodal checkpoint model with the language-only model that performs best in BLiMP (100M checkpoint) with varying weights. Figure~\ref{fig:mergeblimp} illustrates the model-merging results when combining the language-only and multimodal models using $\alpha =$ 0.3, 0.5 and 0.8 as the weights of the language-only model and $1-\alpha$ for the multimodal model. Merging the trained language-only model in the early training stages of the multimodal model meaningfully helps in getting better results in BLiMP. Additionally, in the later checkpoints when the multimodal model's language-only capabilities begin to drop, 0.3/0.7 merging scheme helps the model maintain language-only capabilities.

To check whether merging with the language-only model affects multimodal performance, we also look at the accuracy on the Winoground benchmark after merging models. 
The results for Winoground are provided in Figure~\ref{fig:wgmerged}, showing that in some checkpoints, merging can actually be beneficial without significantly decreasing multimodal scores.

\section{Conclusion}
We have investigated whether model modification in the form of model merging at inference time would benefit multimodal BabyLM models in language-only and multimodal tasks. Our results showed that, indeed, multimodal models tend to underperform in text-only benchmarks that focus on grammar (although surpassing previous baselines) and model merging with text-only models can help alleviate this issue to some extent. Future work can explore other model merging techniques and the effects of model merging in other benchmarks. 

\section*{Limitations}
Due to time and compute constraints, we altered our intended initial setup where the model is fed 256 image patches into one where a single, pooled image representation is relayed to the model. This might cause information loss and performance drop, and ideally, we would like to provide the whole set of image patches. We tested models with 6 transformer layers, which is quite few compared to state-of-the-art models. Therefore, this might have resulted in lesser performance. However, we believe that our results shed light on what to expect in compute and data-efficient/scarce setups, which should be investigated further using more seeds and different training orders for robustness and generalizability of the conclusions. Additionally, although the set of benchmarks we tested on does not cover the full spectrum of language-only and multimodal tasks in the BabyLM challenge, we think that they span a reasonable range of them, providing insights into the dynamics visuo-linguistic processes as training progresses.

\section*{Acknowledgments} 
The research reported in this paper was supported by the European Research Council (ERC), grant 101088098 - MEMLANG. The first author utilized compute resources granted by the Dutch Research Council (NWO). The second author acknowledges support from NWO, Open Competition XS project number 406.XS.25.01.104. Views and opinions expressed are, however, those of the authors only and do not necessarily reflect those of the European Union or the European Research Council Executive Agency. Neither the European Union nor the granting authority can be held responsible for them.

\bibliography{custom}

\end{document}